\documentclass[letterpaper, 10 pt, conference]{ieeeconf}  %
\usepackage{booktabs}
\usepackage{multirow}
\usepackage{xspace}
\usepackage{graphicx}
\usepackage{amsmath}
\usepackage{subfig}
\usepackage{amssymb}
\usepackage{hyperref}

\IEEEoverridecommandlockouts                              %

 \usepackage{amsmath} 
\newtheorem{definition}{Definition}
\newcommand{\statespace}{\ensuremath{\mathbf X}\xspace}
\newcommand{\obsspace}{\ensuremath{\mathbf Y}\xspace}
\newcommand{\actspace}{\ensuremath{\mathbf U}\xspace}
\newcommand{\trajectory}{\ensuremath{z}\xspace}
\newcommand{\initstate}{\ensuremath{x_0}\xspace}
\newcommand{\controller}{\ensuremath{h}\xspace}
\newcommand{\controlleradv}{\ensuremath{h_{\text{adv}}}\xspace}
\newcommand{\dynmodel}{\ensuremath{f}\xspace}
\newcommand{\obsmodel}{\ensuremath{o}\xspace}
\newcommand{\repairmodel}{\ensuremath{r}\xspace}

\title{\LARGE \bf
Generalizable Image Repair for Robust Visual Control
}

\author{Carson Sobolewski$^{1}$, Zhenjiang Mao$^{1}$, Kshitij Maruti Vejre$^{1}$, and Ivan Ruchkin$^{1}$%
\thanks{The authors thank Adam McAleer and Jianing Yu for instrumenting the simulator, and Bilhani Charlakorla for training racing controllers. 
This work is supported in part by the University of Florida AI Scholars Program and the National Science Foundation (NSF) Grants CCF 2403616 and CNS 2513076. Any opinions, findings, and conclusions or recommendations
expressed in this material are those of the author(s) and do not
necessarily reflect the views of the NSF.}
\thanks{$^{1}$%
Trustworthy Engineered Autonomy (TEA) Lab, Department of Electrical and Computer Engineering, University of Florida,
        {\tt\small \{csobolewski, z.mao, kvejre, iruchkin\}@ufl.edu}}%
}

\begin{document}

\maketitle
\thispagestyle{empty}
\pagestyle{empty}

\begin{abstract}
Vision-based control relies on accurate perception to achieve robustness. However, image distribution changes caused by sensor noise, adverse weather, and dynamic lighting can degrade perception, leading to suboptimal control decisions. Existing approaches, including domain adaptation and adversarial training, improve robustness but struggle to generalize to unseen corruptions while introducing computational overhead. To address this challenge, we propose a real-time image repair module that restores corrupted images before they are used by the controller. Our method leverages generative adversarial models, specifically CycleGAN and pix2pix, for image repair. CycleGAN enables unpaired image-to-image translation to adapt to novel corruptions, while pix2pix exploits paired image data when available to improve the quality. To ensure alignment with control performance, we introduce a control-focused loss function that prioritizes perceptual consistency in repaired images. We evaluated our method in a simulated autonomous racing environment with various visual corruptions. The results show that our approach significantly improves performance compared to baselines, mitigating distribution shift and enhancing controller reliability.

Source code: %
\href{https://github.com/Trustworthy-Engineered-Autonomy-Lab/generalizable-image-repair}{github.com/trustworthy-engineered-autonomy-lab/generalizable-image-repair}%

Demonstration video: 
\href{https://youtu.be/C3-WlZpYBm8}{youtu.be/C3-WlZpYBm8}

\end{abstract}

\section{INTRODUCTION}

Vision-based control has achieved significant progress, particularly in racing applications~\cite{cai2021vision}. Despite these advancements, real-world deployment remains challenging due to image distribution shifts~\cite{bauchwitz2022effects} caused by sensor noise, adverse weather conditions (such as rain, fog, and snow), rapid lighting changes, and adversarial perturbations. Such factors can lead to severe image corruption, compromising the stability and reliability of visual control systems. In practice, most existing vision-based methods assume that the training data accurately reflects the test environment. However, this assumption often fails, resulting in subpar decision-making and degraded performance.

Several strategies have been proposed to mitigate the impact of distribution shifts. Domain adaptation~\cite{li2023domain} and generalization techniques~\cite{niemeijer2022domain} can align feature distributions between the training and testing domains, while adversarial training~\cite{goodfellow2014explaining} seeks to improve model robustness by exposing the system to corrupted inputs during training. Although these techniques can be useful, their effectiveness is often constrained by the limited diversity of corruptions seen during training. In real-world scenarios, unforeseen environmental factors --- such as novel sensor noise patterns, extreme weather conditions, or complex lighting variations --- may introduce corruptions that differ significantly from those encountered during training. As a result, these methods struggle to generalize to unseen data, incur high computational costs, or sacrifice accuracy on clean images.
 Furthermore, recent advances in generative vision models have opened new avenues for image synthesis and repair. Models such as generative adversarial networks (GANs)~\cite{goodfellow2020generative}, variational autoencoders (VAEs)~\cite{kingma2013auto}, and denoising diffusion~\cite{ho2020denoising} have shown excellent performance in generating high-quality images, yet their potential for real-time image restoration in robotics remains under-explored.

This paper addresses the problem of image repair for robust control. We introduce a \emph{real-time image repair module} that seamlessly integrates into vision-based control systems. This module repairs degraded images at test time without any prior knowledge of them, thereby reducing the distribution gap between the training and testing environments. %
Our method leverages recent advances in GANs, specifically \textit{CycleGAN}~\cite{zhu2017unpaired} and \textit{pix2pix}~\cite{isola2017image}, to perform real-time image restoration. CycleGAN enables unpaired image-to-image translation, allowing adaptation to unseen corruptions without requiring paired training data, while pix2pix utilizes paired data to learn a direct mapping between corrupted and clean images, leading to more precise restorations when such data is available. 

By utilizing these image translation techniques, our approach balances computational efficiency and repair quality, ensuring minimal latency while preserving essential perception features for reliable control. %
To train our module, we introduce a \emph{control-focused loss function} that ensures the repaired images lead to accurate control outputs, directly improving performance under adverse conditions.

The contributions of our work are summarized below:
\begin{itemize}
 \item A real-time image repair module that integrates into vision-based control systems, enhancing robustness against image corruption.
    \looseness=-1
    \item A test-time application of advanced generative adversarial networks to reduce the distribution shift in vision-based control via a specialized loss function. %

\item A comprehensive evaluation in simulated autonomous racing that simulates five diverse conditions, including adverse weather, lighting variations, and sensor noise. The results demonstrate that our method significantly improves control performance over existing techniques.
\end{itemize}

The remainder of this paper is organized as follows. Section~\ref{sec:problem} defines the problem statement. Section~\ref{sec:approach} describes our method in detail, including the integration of generative models into the control pipeline for real-time image repair.
 Section~\ref{sec:experiment} presents the experimental setup and evaluation results that demonstrate the effectiveness of our approach. In Section~\ref{sec:related}, we review related work in domain adaptation, adversarial training, and generative image repair techniques. Finally, Section~\ref{sec:conclusion} concludes the paper with a discussion of our findings and potential directions for future research.

\begin{figure}
\vspace{3mm}
    \centering
    \subfloat[Normal]{\includegraphics[width=0.4\linewidth]{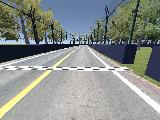}}\hspace{10pt}%
    \subfloat[Darkened]{\includegraphics[width=0.4\linewidth]{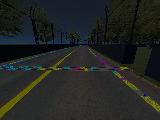}}
    
    \subfloat[Salt and Pepper]{\includegraphics[width=0.4\linewidth]{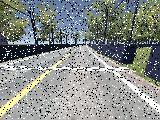}}\hspace{10pt}%
    \subfloat[Rain]{\includegraphics[width=0.4\linewidth]{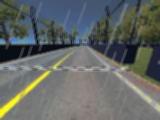}}
    
    \subfloat[Fog]{\includegraphics[width=0.4\linewidth]{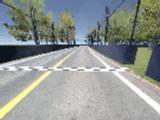}}\hspace{10pt}%
    \subfloat[Snow]{\includegraphics[width=0.4\linewidth]{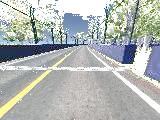}}

    \caption{Sample images from each corrupted distribution. %
    }
    \label{fig:samples}
\end{figure}
\section{PROBLEM STATEMENT}
\label{sec:problem}
\begin{definition}[Vision-based System]
A vision-based system $s=(\statespace,\obsspace,\actspace,\initstate,\controller,\dynmodel,\obsmodel)$ consists of the following:
\begin{itemize}
    \item \textit{State space} \statespace, containing continuous states $x$
	\item \textit{Image observation space} \obsspace, containing images $y$
    \item \textit{Control action space} \actspace, containing control actions $u$
     \item \textit{Initial state} \initstate, from which the system starts executing
     \item \textit{Vision-based control module} $\controller: \obsspace \rightarrow \actspace$, which takes an image as input and is realized by a neural network
     \item \textit{Dynamical model} $\dynmodel: \statespace\times \actspace\rightarrow \statespace$, which describes the system's execution %
     (unknown to us)
     \item \textit{Observation model} $\obsmodel: \statespace\rightarrow \obsspace$, which generates an image observation based on the state (unknown to us)
\end{itemize}

\end{definition}

This paper focuses on the setting of autonomous car racing. Here, the states \statespace include the car's pose and translational and angular velocities, the observations $\obsspace$ are first-person forward-facing images from a camera mounted on the car's chassis, and the actions $\actspace$ are pairs of a steering angle and a throttle command produced by a convolutional neural network (CNN) controller \controller. The dynamical model \dynmodel, observation model \obsmodel, and ground-truth state $x$ are unknown both during operation and at design time.
Once the controller is deployed in some initial state \initstate, the system executes a \textit{trajectory} \trajectory, which is a state-observation-action sequence $\{x_i,y_i,u_i\}^T_{i=0}$ up to some time $T$, where:\begin{equation}
x_{i+1}=\dynmodel(x_i,y_i),\quad y_i=o(x_i),\quad u_i=h(y_i)
\end{equation}

Given a training dataset $\mathcal{D}_{train}$ of paired images and control actions, where the images are drawn i.i.d from the image distribution $P_{train}$ and the actions follow the center line of the track, we define an image-based neural network controller \controller, which translates image observations $y$ into control actions $u$. During operation, we have a testing image distribution $P_{test}$, where images may undergo corruptions and disturbances. 
To formalize the relationship between the testing and training images, we assume that test images in $P_{test}$ are generated from clean images in $P_{train}$ via an unknown random noise $\eta$ from an unknown distribution $P_\eta$. In our experiments, the corruptions we consider include darkness, salt-and-pepper noise, rain, fog, and snow, which we denote as $\eta \in \{\eta_{\text{dark}}, \eta_{\text{salt}}, \eta_{\text{rain}}, \eta_{\text{fog}}, \eta_{\text{snow}}\}$. Examples of these corruptions are shown in Figure~\ref{fig:samples}. Then we consider two settings of image generation:
\begin{itemize}
    \item \textbf{Paired setting:} Each test image $\hat{y} \sim P_{test}$ has a corresponding clean image $y \sim P_{train}$, and we model the corruption $c$ as an %
    image transformation:
    \begin{equation}
        \hat{y} = c(y, \eta), \quad \eta \sim P_{\eta}
    \end{equation}
    where $c$ is an unknown corruption function, which could be additive (e.g., $\hat{y} = y + \eta$ for noise perturbations) or more complex (e.g., contrast reduction, motion blur). This setting is suitable for conditional image generation approaches such as pix2pix.
    
    \item \textbf{Unpaired setting:} There is no one-to-one mapping between test and clean images, and we assume that $P_{test}$ is a corrupted version of $P_{train}$ in a distributional sense. Instead of a direct transformation, we model:
    \begin{equation}
        P_{test} = C(P_{train}, \eta), \quad \eta \sim P_{\eta}
    \end{equation}
    where $C$ represents an unknown transformation influenced by noise distribution $P_{\eta}$. This setting is suitable for domain adaptation approaches such as CycleGAN.
\end{itemize}

\begin{figure}
    \centering
    \includegraphics[width=\linewidth]{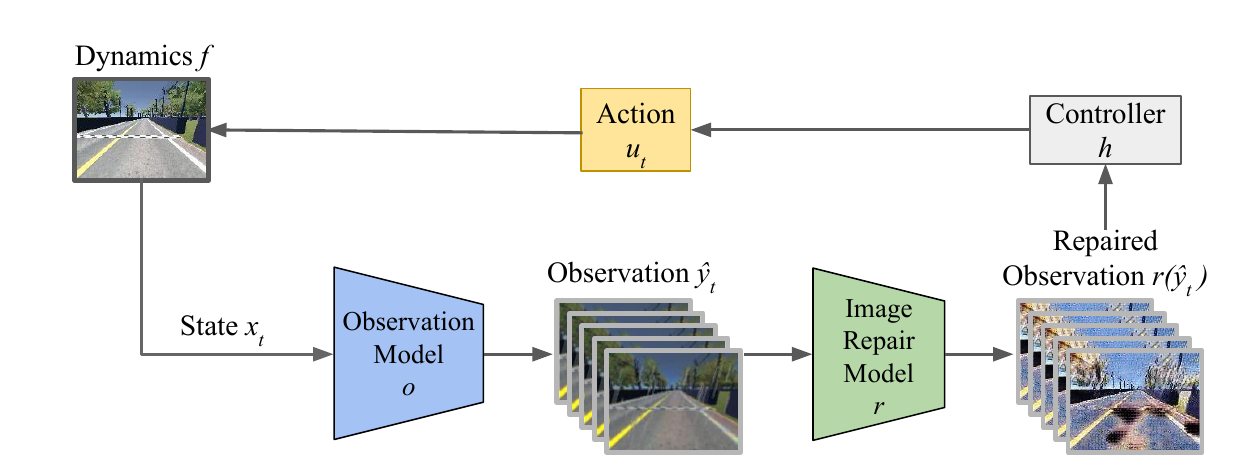}
    \caption{
    Our approach architecture: an image repair model repairs images before they are used for control.
    }
    \label{fig:architecture}
\end{figure}

\looseness=-1
We define an \textit{image repair model}, $\repairmodel:\obsspace\rightarrow \obsspace$, that translates images from the testing domain to the training domain \textit{without} explicit knowledge of the corruption functions $c$ and $C$ or noises $\eta$ or $P_\eta$. Intuitively, the repair model $\repairmodel$ aims to shift a corrupted image $\hat{y}$ closer to training distribution $P_{train}$ before passing it to the controller:
\begin{equation}
    y' = \repairmodel(\hat{y}),
\end{equation}
where $y'$ is the repaired image. Our architecture is shown in Figure~\ref{fig:architecture}. Images from the observation model, $o$, are passed to our repair model $\repairmodel$, to obtain repaired images. These repaired images are then used by the controller $h$ to predict control actions.

Ultimately, we aim to find a repair model $\repairmodel$ that minimizes the difference between the system trajectories taken by our controller on these images: 
\begin{equation}\label{eq:traj}
    \min_{\repairmodel} \| z_{\controller(y)} - z_{\controller(\repairmodel(\hat{y}))} \|_2, 
\end{equation}
where $y$ is an in-distribution image from $P_{train}$, $\hat{y}$ is an out-of-distribution image from $P_{test}$, and $z_i$ is the trajectory generated from the control actions $u$ produced by setup $i$. 

For the autonomous car racing setting, we calculate this difference in trajectory by comparing mean-squared cross-track error (CTE) values with respect to the center line of the track. 

\section{IMAGE REPAIR FOR CONTROL}
\label{sec:approach}

This section introduces the repair techniques that we experimented with and then describes our control-focused training for generative models. 

\looseness=-1
\subsection{Repair Techniques}\label{sec:configs}
We implement three categories of approaches to combat distribution shift from corrupted images. One category is to conduct \textit{adversarial training}, to generalize the controller over some corruptions. An alternative category is \textit{adversarial restoration} --- directly removing the perturbations from the images with computer vision techniques. Finally, \textit{generative models} are capable of repairing images with limited (if any) knowledge of distribution $P_\eta$ of their corruptions. We implement all three of the aforementioned approaches and investigate their generalization to unseen corruptions.

\smallskip
\noindent
\textbf{Adversarial training}: Many previous works in adversarial training opt for learning-based methods to repair adversarial images. For instance, FGSM~\cite{goodfellow2014explaining}, PGD~\cite{madry2017towards}, and TRADES~\cite{zhang2019theoretically} use supervised learning, \cite{pinto2017robust} uses reinforcement learning, and \cite{ganin2016domain} uses domain adaptation. While these methods have relatively low computational complexity, we instead opt to use vision-based disturbances for simplicity. We train the controller model $\controlleradv$ on a combination of normal images and \textit{three} types of corrupted images, optimizing it to generalize to unseen noise types. Specifically, the adversarial training objective minimizes the discrepancy between control actions on clean and corrupted images:
\begin{equation}
    \min_{\controlleradv} \mathop{\mathbb{E}}_{\substack{y \sim P_{train},\\ \hat{y} \sim P_{corrupt}}} \| \controlleradv(y) - \controlleradv(\hat{y}) \|_2^2,
\end{equation}
where $P_{corrupt}$ represents the distribution of images corrupted by known perturbations. This ensures that $\controlleradv$ learns to make consistent control predictions despite variations in image quality. The trained controller is then evaluated on \textit{two} unseen corruption types to assess its generalization ability.
 Through this method we aim to include common corruptions in the training data, increasing the likelihood that the data from $P_{test}$ aligns with $P_{train}$.

\smallskip
\noindent
\textbf{Adversarial restoration}: Much like in adversarial training, many restoration methods opt for learning-based approaches. For instance, PixelDefend~\cite{song2017pixeldefend} and generative adversarial denoising \cite{samangouei2018defense} reconstruct perturbed images to restore clean data representations, randomized smoothing \cite{cohen2019certified} certifies robustness by averaging predictions over noisy variations of an input, and feature denoising networks \cite{xie2019feature} integrate adversarial noise suppression within convolutional architectures. However, these methods also come with increased computational complexity, and their efficacy often depends on the strength and nature of adversarial attacks. 

Instead, we opt for two computer vision approaches: \textit{Lucy Richardson Deconvolution}~\cite{richardson1972bayesian}, and a \textit{Variational Bayes}~\cite{tzikas2009variational}. Lucy Richardson Deconvolution is an iterative maximum-likelihood method for removing blur caused by a known or estimated point spread function (PSF), commonly used in optical and motion deblurring. The Variational Bayesian approach, on the other hand, uses probabilistic inference to estimate the clean image distribution by minimizing the divergence between an approximate posterior and the true posterior, making it effective for denoising under complex noise models. We prepend these computer vision techniques to the controller, repairing images before the controller predicts control actions. Formally, we define both vision-based restoration approaches as follows:
\begin{equation}
    r_{\{\text{LR},\text{VB}\}}(\hat{y})=y',
\end{equation}
where $r_{\text{LR}}$ is the Lucy Richardson Deconvolution approach, $r_{\text{VB}}$ is the Variational Bayesian approach, $\hat{y}$ is the original image from $P_{test}$, and $y'$ is the repaired image that is then passed to the controller as follows:
\begin{equation}
    u =h(y'),
\end{equation}
where $u$ is the action of controller $h$ on repaired image $y'$.

\smallskip
\noindent
\textbf{Generative models}: Generative models, such as GANs~\cite{goodfellow2020generative}, VAEs~\cite{kingma2013auto}, autoregressive models~\cite{van2016pixel}, and diffusion models~\cite{ho2020denoising}, are capable of repairing images with limited knowledge of the corrupting distribution $P_\eta$. In particular, GAN-based approaches, including StyleGAN~\cite{karras2019style}, CycleGAN~\cite{zhu2017unpaired}, BigGAN~\cite{brock2018large}, and pix2pix~\cite{isola2017image} have demonstrated high-quality image synthesis by leveraging adversarial training. VAEs focus on learning a structured latent space for controllable generation. Diffusion models like DDPM~\cite{ho2020denoising} and Stable Diffusion~\cite{rombach2022high} have achieved state-of-the-art results in text-to-image generation and high-resolution image synthesis --- but come at the cost of high computational complexity and long inference times that inhibit real-time usage. 

\looseness=-1
We implement three generative approaches --- VAEs, CycleGAN, and pix2pix --- and compare their performance in Section~\ref{sec:experiment}. We prepend our generative repair model \repairmodel to \controller, repairing corrupted images $\hat{y}\in P_{test}$ and passing them to $h$:
\begin{equation}\label{eq:gen}
    r_{\{\text{VAE},\text{CG},\text{P2P}\}}(\hat{y})=y', \quad u = h(y')
\end{equation}
where $r_{\text{VAE}}$ is a VAE repair model, $r_{\text{CG}}$ is a CycleGAN repair model, $r_{\text{P2P}}$ is a pix2pix repair model, %
and $y'$ is the resulting repaired image.

\subsection{Generative Model Training}\label{sec:training}

To implement the VAE, we train it using dataset $\mathcal{D}_{train}$ drawn from $P_{train}$ consisting of in-distribution images. Then, by encoding and decoding the image, we will get images $y'$ that are closer to the training distribution $P_{train}$.

\begin{figure}
\vspace{3mm}
    \centering
    \subfloat[Forward Direction $te\to tr$]{\includegraphics[width=0.99\linewidth]{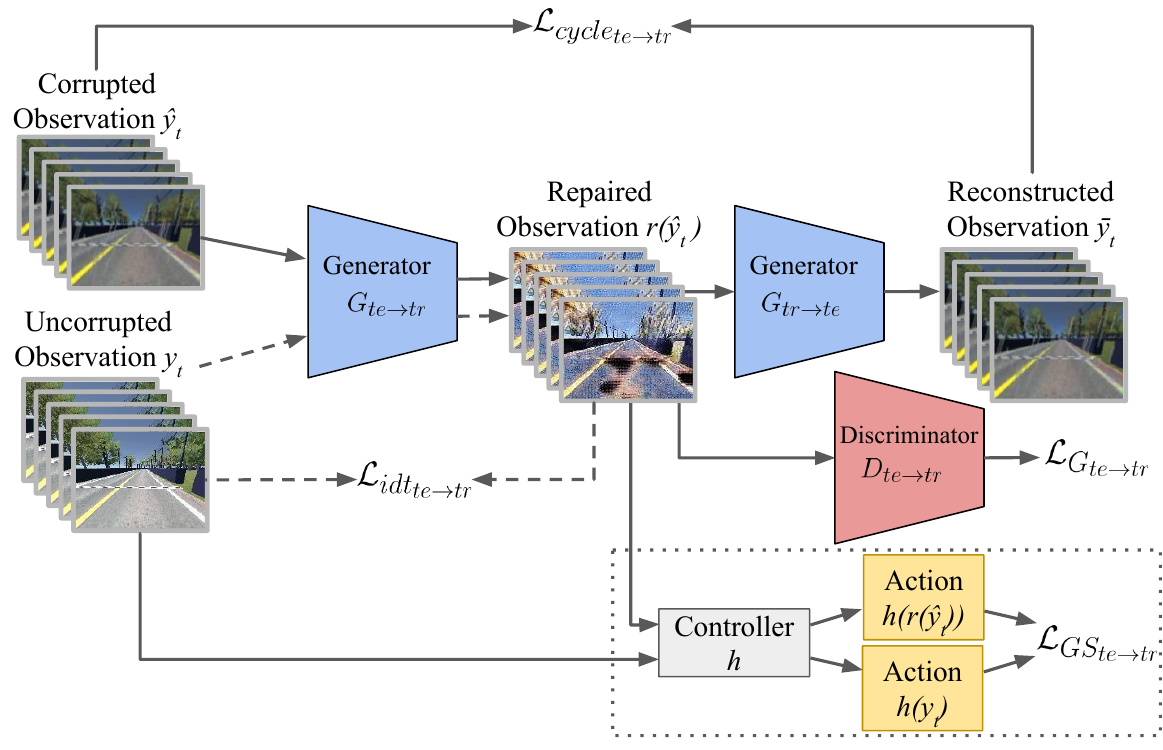}}
    
    \subfloat[Backward Direction $tr\to te$]{\includegraphics[width=0.99\linewidth]{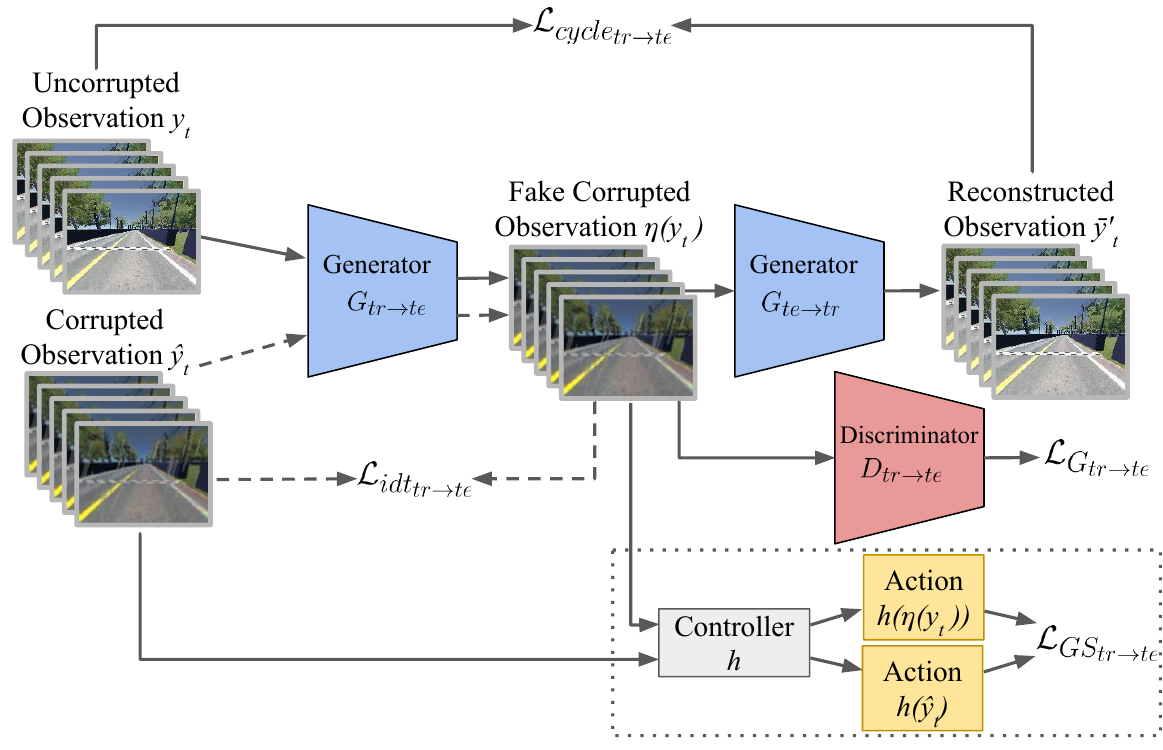}}

    \caption{Data flow for the forward and backward training passes of our CycleGAN's generators. Dashed lines are for identity loss $\mathcal{L}_{idt}$, whereas solid lines are for the other losses. The dotted box contains the flow for our controller loss.
    }
    \label{fig:cyclegan}
\end{figure}

For CycleGAN, we train it in two ways: with and without a controller loss. Without controller loss, the CycleGAN is trained to translate between image distributions $P_{test}$ (which consists of all corrupted images) and $P_{train}$ (which consists of clean images) in both directions. CycleGAN does not require paired data but instead learns a bidirectional mapping between these two distributions. To do so, CycleGAN takes images from both of the aforementioned distributions, doing cyclical passes in both directions --- repairing images and re-corrupting them and then corrupting images and repairing them. This bidirectional training process is depicted in Figure~\ref{fig:cyclegan}. We train for 100 epochs and then take generator $G_{te \to tr}$ (which translates images from $P_{test}$ to $P_{train}$) as our image repair model.

Since we ultimately aim to train this repair model such that it minimizes the performance impacts of image corruptions (see Equation \ref{eq:traj}), we introduce a (novel to our knowledge) way to train the CycleGAN with the controller in the loop. Unfortunately, we cannot minimize the trajectory difference directly per Equation~\ref{eq:traj} --- and instead opt to minimize the difference in control actions between the controller on a normal image and the repaired image (in the paired setting):
\begin{equation}
    \min_{\repairmodel} \mathop{\mathbb{E}}_{\substack{y \sim P_{train},\\ \hat{y} \sim P_{test}}} \left[ \| \controller(y) - \controller(\repairmodel(\hat{y})) \| \right],
\end{equation}
where $y$ is an in-distribution image from $P_{train}$, $\hat{y}$ is an out-of-distribution image from $P_{test}$, $r$ is the CycleGAN generator $G_{te\to tr}$, and $h$ is the controller for which we are repairing images. However, this comes at the cost of requiring a paired dataset. 

We also implement additional losses $\mathcal{L}_{GS_{te\to tr}}$ and $\mathcal{L}_{GS_{tr\to te}}$, representing the L1 loss (mean absolute error) between the predicted control actions on normal images $y$ and predicted images $y'$ from generators $G_{te\to tr}$ and $G_{tr\to te}$, respectively. These losses are then added to the main CycleGAN generator loss computation as follows:
\begin{equation}
\begin{split}&\mathcal{L}_{CG}=\mathcal{L}_{G_{te\to tr}}+\mathcal{L}_{G_{tr\to te}} \\
    &+\lambda_1\mathcal{L}_{cycle_{te\to tr}} +\lambda_2\mathcal{L}_{cycle_{tr\to te}} \\
    &+\lambda_3(\lambda_2\mathcal{L}_{idt_{te\to tr}}+\lambda_1\mathcal{L}_{idt_{tr\to te}}) \\
    &+\lambda_4(\mathcal{L}_{GS_{te\to tr}}+\mathcal{L}_{GS_{tr\to te}}),
\end{split}
\end{equation}
where $\mathcal{L}_{G_{te\to tr}}$ and $\mathcal{L}_{G_{tr\to te}}$ are standard GAN adversarial losses, $\mathcal{L}_{cycle_{te\to tr}}$ and $\mathcal{L}_{cycle_{tr\to te}}$ enforce cycle consistency, and $\mathcal{L}_{idt_{te\to tr}}$ and $\mathcal{L}_{idt_{tr\to te}}$ discourage the model from modifying target-domain images. The generator training flow is illustrated in Figure~\ref{fig:cyclegan}.  Hyperparameters $\lambda_1 \dots \lambda_4$ modulate the effect of existing CycleGAN losses and our controller losses $\mathcal{L}_{GS_{te\to tr}}$ and $\mathcal{L}_{GS_{tr\to te}}$.

For pix2pix, the training parallels that of CycleGAN, but with the requirement of having a paired dataset of normal and corrupted images. We train for 30 epochs and then take the generator as our repair model. Much like for CycleGAN, we implement an additional controller loss $\mathcal{L}_{GS}$, representing the L1 loss between the predicted control action on normal images $y$ and predicted images $y'$ from the generator. The loss is then added to the existing pix2pix generator loss:
\begin{equation}
    \mathcal{L}_{PG}=\mathcal{L}_{GAN}+\lambda_5\mathcal{L}_{L1}+\lambda_6\mathcal{L}_{GS},
\end{equation}
where $\mathcal{L}_{GAN}$ is a typical GAN adversarial loss, $\mathcal{L}_{L1}$ is the L1 loss between the pixel values of the generated and original images, and $\lambda_5$ and $\lambda_6$ are hyperparameter values to modulate the effect of the L1 and controller losses. %

\section{EXPERIMENTAL EVALUATION}\label{sec:experiment}

\begin{table*}[th]
\vspace{3mm}
    \centering
    \scriptsize 
    \caption{Racing Performance: Root Mean-Squared CTE (m, $\downarrow$) for Different Configurations Across Corruption Types}
    \label{tab:cte_values}
    \begin{tabular}{llllcccccc|cc}
        \toprule
        Config & Approach & Paired-ness & Controller Type & Normal & Brightness & Salt/Pepper & Rain & Fog & Snow & \textbf{Unseen} & \textbf{All} \\
        \midrule
        - & Original Controller $h$ & - & - & 1.545 & 3.062 & 1.571 & 2.257 & 1.135 & 2.118 & 2.131 & 2.045 \\
        \midrule
        - & Lucy Richardson & - & - & 2.957 & 3.148 & 2.459 & 2.088 & 1.969 & 1.923 & 2.362 & 2.471 \\
        - & Variational Bayes & - & - & 2.158 & 3.078 & 2.552 & 2.733 & 2.523 & 2.275 & 2.646 & 2.571 \\
        \midrule
        \multirow{8}{*}{A} & Generalized Controller $h_{adv}$ & - & - & 1.410 & 1.627 & 1.469 & 1.520 & 2.338 & 2.250 & 2.295 & 1.810 \\
        & VAE & - & - & 1.395 & 3.147 & 2.088 & 1.838 & 1.942 & 2.220 & 2.085 & 2.171 \\
        & CycleGAN & Unpaired & tr: $\varnothing$, te: $h$ & 1.611 & 1.499 & 2.559 & 1.424 & 1.178 & 1.965 & 1.620 & 1.764 \\
        & CycleGAN & Unpaired & tr: $\varnothing$, te: $h_{adv}$ & 1.282 & 1.525 & 1.726 & 1.533 & 1.475 & 2.306 & 1.936 & 1.673 \\
        & CycleGAN & Paired & tr: $h$, te: $h$ & \textbf{1.242} & \textbf{1.353} & \textbf{1.345} & \textbf{1.392} & \textbf{1.087} & \textbf{1.833} & \textbf{1.507} & \textbf{1.394} \\
        & CycleGAN & Paired & tr: $h$, te: $h_{adv}$ & 1.519 & 1.392 & 2.234 & 1.396 & 2.347 & 2.364 & 2.355 & 1.927 \\
        & CycleGAN & Paired & tr: $h_{adv}$, te: $h$ & 1.466 & 1.709 & 1.885 & 1.405 & 2.069 & 1.956 & 2.013 & 1.765 \\
        & CycleGAN & Paired & tr: $h_{adv}$, te: $h_{adv}$ & 1.502 & 1.763 & 1.602 & 1.663 & 1.480 & 2.320 & 1.946 & 1.745 \\
        & pix2pix & Paired & tr: $\varnothing$, te: $h$ & 1.648 & 1.403 & 1.552 & 1.451 & 1.415 & 2.003 & 1.734 & 1.605 \\
        & pix2pix & Paired & tr: $\varnothing$, te: $h_{adv}$ & 1.588 & 1.767 & 1.628 & 2.323 & 1.464 & 2.154 & 1.842 & 1.847 \\
        & pix2pix & Paired & tr: $h$, te: $h$ & 1.984 & 1.950 & 1.966 & 1.922 & 1.602 & 2.028 & 1.936 & 1.951 \\
        & pix2pix & Paired & tr: $h$, te: $h_{adv}$ & 1.635 & 1.552 & 2.261 & 2.312 & 2.361 & 2.276 & 2.319 & 2.095 \\
        & pix2pix & Paired & tr: $h_{adv}$, te: $h$ & 1.817 & 1.846 & 1.931 & 1.997 & 2.121 & 1.959 & 2.042 & 1.948 \\
        & pix2pix & Paired & tr: $h_{adv}$, te: $h_{adv}$ & 1.655 & 2.189 & 1.534 & 2.149 & 2.254 & 2.124 & 2.190 & 2.004 \\
        \midrule
        \multirow{8}{*}{B} & Generalized Controller $h_{adv}$ & - & - & 1.549 & 1.734 & 1.631 & 1.582 & 1.396 & 2.119 & 1.870 & 1.684 \\
        & VAE & - & - & 1.393 & 3.147 & 2.146 & 1.923 & 1.935 & 2.071 & 1.998 & 2.167 \\
        & CycleGAN & Unpaired & tr: $\varnothing$, te: $h$ & 1.500 & 1.446 & 1.723 & \textbf{1.264} & \textbf{1.266} & 2.189 & 1.788 & 1.597 \\
        & CycleGAN & Unpaired & tr: $\varnothing$, te: $h_{adv}$ & 1.405 & \textbf{1.419} & 1.516 & 1.639 & 2.239 & 2.106 & 1.887 & 1.752 \\
        & CycleGAN & Paired & tr: $h$, te: $h$ & 1.570 & 1.506 & \textbf{1.117} & 1.513 & 1.338 & 1.995 & 1.770 & \textbf{1.530} \\
        & CycleGAN & Paired & tr: $h$, te: $h_{adv}$ & 1.507 & 1.567 & 1.412 & 1.668 & 1.639 & 2.075 & 1.883 & 1.658 \\
        & CycleGAN & Paired & tr: $h_{adv}$, te: $h$ & \textbf{1.265} & 2.332 & 2.782 & 2.580 & 2.830 & 1.869 & 2.253 & 2.343 \\
        & CycleGAN & Paired & tr: $h_{adv}$, te: $h_{adv}$ & 1.455 & 2.628 & 2.265 & 2.297 & 2.702 & 2.166 & 2.233 & 2.289 \\
        & pix2pix & Paired & tr: $\varnothing$, te: $h$ & 2.652 & 2.250 & 1.853 & 1.586 & 1.673 & 1.865 & 1.731 & 2.013 \\
        & pix2pix & Paired & tr: $\varnothing$, te: $h_{adv}$ & 1.594 & 1.719 & 1.676 & 1.888 & 2.124 & 2.071 & 1.981 & 1.856 \\
        & pix2pix & Paired & tr: $h$, te: $h$ & 1.840 & 1.548 & 1.540 & 1.529 & 1.454 & 1.919 & 1.784 & 1.656 \\
        & pix2pix & Paired & tr: $h$, te: $h_{adv}$ & 2.063 & 1.614 & 1.643 & 1.809 & 1.761 & 2.085 & 1.952 & 1.838 \\
        & pix2pix & Paired & tr: $h_{adv}$, te: $h$ & 1.796 & 1.804 & 1.682 & 1.574 & 1.406 & \textbf{1.341} & \textbf{1.523} & 1.643 \\
        & pix2pix & Paired & tr: $h_{adv}$, te: $h_{adv}$ & 1.801 & 1.875 & 1.743 & 1.817 & 1.621 & 2.066 & 1.946 & 1.826 \\
        \bottomrule
    \end{tabular}
    \vspace{-4mm}
\end{table*}

\begin{table}[h]
    \centering
    \caption{Average Inference Time per Image}
    \footnotesize
    \label{tab:tab2}

    \begin{tabular}{lc}
        \toprule
        \textbf{Model} & \textbf{Time (ms)} \\
        \midrule
        Original Controller ($h$) &  1.76 $\pm$ 0.84 \\
        Lucy Richardson & 0.96 $\pm$ 0.20
\\
        Variational Bayes & 5.38 $\pm$ 1.49
\\
        VAE & 2.52 $\pm$ 3.57
\\
        CycleGAN &  23.86 $\pm$ 6.40
\\
        pix2pix & 6.88 $\pm$ 2.47
 \\
        \bottomrule
    \end{tabular}
\end{table}

\noindent
\textbf{Experimental Setup.} The primary objective is to assess how well different repair techniques restore corrupted images and improve control performance under various disturbances.
To evaluate the effectiveness of our image repair methods in vision-based autonomous racing, we conduct experiments in a simulated environment for DonkeyCar racing (see \url{https://www.donkeycar.com/}). Simulation is chosen over real-world deployment due to its controlled setting, reproducibility, ability to systematically introduce diverse corruptions, and an easy-to-quantify autonomous racing task. 
On the other hand, open datasets like nuScenes and Waymo Open do not allow counterfactual executions under visual corruptions, whereas implementing repeatable paired corruptions in a physical setting can be very effortful. The implementation of our evaluation can be found in an online repository: \url{https://github.com/Trustworthy-Engineered-Autonomy-Lab/generalizable-image-repair}. %

\smallskip
\noindent
\textbf{Techniques under Comparison.} We compare our proposed repair methods against multiple baselines. The first baseline is the original controller --- a CNN trained only on clean images from $P_{train}$, without exposure to corrupted inputs. Another baseline is the \textit{generalized controller}, which is adversarially trained on a mix of clean images and three types of corruptions, aiming to generalize to unseen corruptions. 
Additionally, we include two computer vision baselines that apply traditional vision-based restoration techniques. We also evaluate a VAE-based repair model, where a VAE is trained to denoise corrupted images before they are used for control. %

We propose using CycleGAN and pix2pix. For CycleGAN, we experiment with both unpaired and paired training with a controller loss. The unpaired CycleGAN learns a mapping between $P_{test}$ (corrupted images) and $P_{train}$ (clean images) without explicit supervision. In contrast, the paired CycleGAN is trained with corrupted-clean image pairs and incorporates an additional loss function to align repaired images with optimal control actions. Finally, we train a pix2pix model on paired corrupted-clean images, both with and without a controller loss, to examine its effectiveness compared to CycleGAN-based methods.

\looseness=-1
\smallskip
\noindent
\textbf{Training Configurations.} To evaluate generalization, two configurations split our five noises $\eta$ (see the end of Section~\ref{sec:problem}) into three in training and two in testing. In \textit{Configuration A}, repair models are trained on normal images, brightness perturbations (0.25 coefficient), salt-and-pepper noise, and rain --- and then tested on snow and fog. In \textit{Configuration B}, repair models are trained on normal images, brightness perturbations, salt-and-pepper noise, and fog --- and tested on rain and snow. These splits were chosen to train on the more basic/synthetic noises and test on the more realistic ones. Each proposed repair model is evaluated with two controllers: the original $h$ trained only on clean images and the generalized $h_{adv}$ exposed to some corruptions in training. While training our CycleGAN models, we use 10 for $\{\lambda_1, \lambda_2, \lambda_4\}$ and 0.5 for $\lambda_3$. For pix2pix, we use 100 for $\{\lambda_5, \lambda_6\}$.

\smallskip
\noindent
\textbf{Evaluation Metrics.} To quantify repair effectiveness and control stability, we use root mean-squared CTE as the primary evaluation metric, which measures deviation from the track center. Additionally, we provide CTE values for each model across unseen corruptions and all corruptions to demonstrate their generalization capabilities.

\smallskip
\noindent
\textbf{Results and Analysis.} Table~\ref{tab:cte_values} presents the performance of each repaired controller in terms of RMSE CTE values across different corruption types under both configurations. For the CycleGAN and pix2pix results, models were trained with no controller loss ($\varnothing$), controller loss on the original controller ($h$), and controller loss on the generalized controller ($h_{adv}$). Then, these models are tested on the original controller ($h$) and the generalized controller ($h_{adv}$). %
As expected, the original controller performs poorly under corruptions, particularly on brightness and rain, highlighting the need for image repair. The generalized controller shows slightly improved performance thanks to domain generalization but struggles with unseen disturbances like fog and snow.

Among the repair methods, traditional computer vision techniques (Lucy Richardson Deconvolution and Variational Bayesian restoration) and the VAE repair model provide limited improvements and fail to generalize effectively across corruptions. In contrast, GAN-based repair models demonstrate substantial gains. Both CycleGAN and pix2pix improve control performance across corruption types. Furthermore, the CycleGAN model trained with our controller loss on the original controller consistently achieves the lowest CTE across nearly all settings, including for unseen corruptions, demonstrating its strong ability to reduce distribution shift while maintaining control performance.

Overall, these results confirm the effectiveness of integrating generative repair models for mitigating visual corruptions. Notably, incorporating a controller loss during CycleGAN training significantly enhances repair quality, leading to more stable and reliable control under challenging conditions. This highlights the importance of aligning image repair objectives with downstream control performance, rather than relying solely on visual reconstruction quality.

\looseness=-1
\smallskip
\noindent
\textbf{Inference Time.} 
To evaluate the run-time feasibility, we report the average inference time and standard deviation in Table~\ref{tab:tab2}. Each model was evaluated on 1000 images from the test set using an NVIDIA L4 GPU. Overall, Lucy Richardson and the original controller exhibit the fastest inference, followed by VAE and Variational Bayes. In contrast, pix2pix and CycleGAN are notably slower yet still feasible, indicating a trade-off between performance and computing cost.

\begin{figure}[th]
\vspace{3mm}
    \centering
    \subfloat[Normal]{\includegraphics[width=0.4\linewidth]{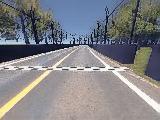}}\hspace{10pt}%
    \subfloat[Darkened]{\includegraphics[width=0.4\linewidth]{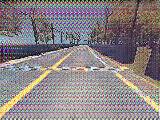}}
    
    \subfloat[Salt and Pepper]{\includegraphics[width=0.4\linewidth]{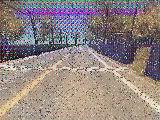}}\hspace{10pt}%
    \subfloat[Rain]{\includegraphics[width=0.4\linewidth]{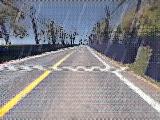}}
    
    \subfloat[Fog]{\includegraphics[width=0.4\linewidth]{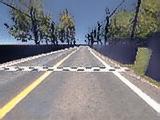}}\hspace{10pt}%
    \subfloat[Snow]{\includegraphics[width=0.4\linewidth]{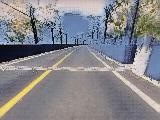}}

    \caption{Sample repaired images for each corruption (and normal) in Figure~\ref{fig:rep_samples} passed through the Configuration A CycleGAN model trained and tested with our controller loss $\mathcal{L}_{CG}$ on the original controller $h$. %
    \vspace{-3mm}
    }
    \label{fig:rep_samples}
\end{figure}

\section{RELATED WORK}
\label{sec:related}

\subsection{Domain Adaptation and Generalization}
Domain adaptation~\cite{farahani2021brief} is a key technique for addressing distribution shift~\cite{wiles2021fine} by adapting a model from in-distribution data to out-of-distribution data. Supervised methods such as fine-tuning~\cite{yosinski2014transferable} and feature alignment~\cite{sun2016return} minimize discrepancies between source and target distributions. Without labeled target data, unsupervised domain adaptation, including data augmentation~\cite{shorten2019survey} and test-time adaptation~\cite{zhang2022memo}, enhances model generalization by using diverse training samples.
Beyond adjusting model parameters, image restoration offers an alternative for mitigating distribution shifts. Diffusion models have been explored for noise removal, effectively filtering away perturbations~\cite{gao2023back}. Instead of shifting the controller to align with a known new distribution (as in domain adaptation), our method enables controller generalization to unseen disturbances by shifting the inputs closer to the initial training distribution. Our approach has the benefit of not requiring images from the target domain and being largely controller-agnostic, especially in configurations without a controller loss.

\looseness=-1
\subsection{Robust Imitation Learning}
Imitation learning (IL)~\cite{hussein2017imitation} enables models to learn control policies from expert demonstrations but is prone to distribution shift. Robust imitation learning (RIL) techniques aim to improve generalization and stability. Methods such as adversarial imitation learning~\cite{rajeswaran2016epopt} leverage adversarial training to align expert and learned policies, while distributionally robust optimization~\cite{santara2017rail} accounts for worst-case domain shifts. Additionally, Bayesian imitation learning~\cite{brown2020bayesian} integrates uncertainty modeling to handle noisy or limited demonstrations, thus enhancing robustness. Similar to these approaches, we conducted adversarial training as a baseline to expose our controller to a wider set of potential settings. By training on some known corruptions alongside normal images, the controller could then generalize better to adversarial samples. However, our generative approaches significantly outperformed the generalized controller: despite the larger training set, there is still the risk that inputs are outside of its generalized training distribution.

\subsection{Adversarial Robustness}

\looseness=-1
For robustness against adversarial perturbations, \textit{adversarial training} (AT)~\cite{wong2020fast} uses adversarial examples in training. However, as previously discussed, AT still struggles to generalize to adversarial examples outside of the training set. While the controller performs well on the three known corruptions, it struggles to generalize to the other two unseen noises.

\looseness=-1
Instead of modifying training, \textit{adversarial purification} removes perturbations from input images. To compare our method to purification approaches, we tested Lucy Richardson Deconvolution and Variational Bayes, which proved detrimental to controller performance on both normal and corrupted images. Thus, classical computer vision approaches struggle to handle complex and unstructured noises, such as those seen in the real world (rain, fog, snow).%

\subsection{Generative Image Repair}

Generative models have been widely adopted for image restoration. Models such as VAEs have reasonable repair capabilities but limited performance compared to more complex models like GANs and diffusion. 
For real-time image repair, efficient GAN variants such as ESRGAN~\cite{wang2018esrgan} and FastGAN~\cite{liu2021fasterstabilizedgantraining} optimize repair speed. Transformer models like SwinIR~\cite{liang2021swinir} enhance denoising through window-based self-attention. Neural filtering techniques~\cite{gharbi2017deep} and self-supervised denoising~\cite{xu2020noisy} provide lightweight alternatives to image restoration. Unfortunately, diffusion models have significant computational complexity, with inference times that are unrealistic for real-time implementation, such as in our case study of autonomous racing. We observed that GANs strike a balance between significant repair capabilities and relatively low computational cost. 
 This is especially the case in the presence of paired data and with the addition of our controller loss that optimizes for repaired images that maximize controller performance.

\section{DISCUSSION AND CONCLUSION}\label{sec:conclusion}

This paper proposed an image repair framework for vision-based autonomous control with enhanced robustness against visual corruptions from sensor noise, adverse weather, and lighting variations. By leveraging advanced generative adversarial models, our system restores corrupted images before control inference, effectively reducing distribution shift between training and testing environments~\cite{ganin2015unsupervised,quinonero2022dataset}. Extensive experiments demonstrate that our approach significantly improves control performance across diverse corruption types. In particular, CycleGAN with a controller-aligned loss function achieves the best overall stability and accuracy, confirming the benefits of integrating control-aware objectives into the repair process~\cite{codevilla2018end}. Our findings suggest that real-time image repair, when coupled with generative modeling and task-specific supervision, is a promising direction for improving the reliability of vision-based autonomous systems under challenging and unpredictable conditions.

While our proposed real-time image repair framework significantly improves control robustness under diverse visual corruptions, several directions remain for future exploration to further enhance system reliability and adaptability. Currently, our system applies image repair to all incoming observations without distinguishing whether an image is corrupted. However, unnecessary repair of clean, in-distribution images can introduce artifacts or latency. To address this, we plan to integrate an OOD detection module that dynamically identifies corrupted or OOD inputs in real time~\cite{lee2018simple}. By coupling OOD detection with repair, the system can selectively trigger the latter only when distribution shifts are detected, thereby minimizing unnecessary computation and preserving original image fidelity when no repair is needed. 

Our current design evaluates repair methods in isolation. However, in practice, different types of corruptions may require distinct restoration strategies. Future work will explore an ensemble repair pipeline to combine multiple repair methods~\cite{ulyanov2018deep,zhang2021plug}. For each incoming corrupted image, the system will compute confidence scores --- such as reconstruction error, perceptual similarity metrics, or control prediction stability --- for outputs from multiple repair models. By selecting the repaired image with the highest confidence, or by adaptively weighting multiple repairs, the system can dynamically choose the most reliable restoration to optimize downstream control performance~\cite{han2022out}.%

Another important direction is to align the controller-based and human-based perspectives on image quality. For example, the high-frequency noise observed in Fig.~\ref{fig:rep_samples} arises because the CNN controller has implicitly learned to rely on high-frequency cues, which can exacerbate overfitting to noise or artifacts~\cite{geirhos2018imagenet,jo2017measuring}. As a consequence, our repair method learns to reproduce these high-frequency features in the repaired images. Future work will mitigate this effect by encouraging the controller to focus on more robust, human-aligned features. For instance, integrating randomized smoothing~\cite{cohen2019certified} into the control pipeline can reduce sensitivity to high-frequency perturbations, thus promoting the use of lower-frequency and more visually pleasing features. By combining such controller designs with repair mechanisms that prioritize perceptual coherence, we hope to achieve a balance between control performance and visual fidelity in vision-based control systems. %

\bibliographystyle{IEEEtran}  
\bibliography{ref}

\end{document}